\documentclass[letterpaper, 10pt, journal, twoside]{IEEEtran}  

\usepackage{graphicx} 
\usepackage{amsmath} 
\usepackage{amssymb}  
\usepackage{xfrac}
\usepackage{gensymb}
\usepackage{subfig}

\begin{document}
\title{Design and Analysis of 6-DOF Triple Scissor Extender Robots with Applications in Aircraft Assembly*}

\markboth{IEEE Robotics and Automation Letters. Preprint Version. Accepted January, 2017}{Gonzalez \MakeLowercase{\textit{et al.}}: Design and Analysis of 6-DOF Triple Scissor Extender Robots}  

\author{Daniel J. Gonzalez$^{1}$ and H. Harry Asada$^{1}$%
	\thanks{Manuscript received: September, 10, 2016; Revised December, 23, 2016; Accepted January, 28, 2017.}
	\thanks{This paper was recommended for publication by Editor Kevin Lynch upon evaluation of the Associate Editor and Reviewers' comments. *This work was supported by The Boeing Company.}
	\thanks{$^{1}$Daniel J. Gonzalez and H. Harry Asada are with the d'Arbeloff Laboratory for Information Systems and Technology in the Department of Mechanical Engineering,
		Massachusetts Institute of Technology, Cambridge, MA 02139, USA.
		{\tt\small \{dgonz, asada\}@mit.edu}}%
	\thanks{Digital Object Identifier (DOI): see top of this page.}
}

\maketitle

\begin{abstract}
A new type of parallel robot mechanism with an extendable structure is presented, and its kinematic properties and design parameters are analyzed. The Triple Scissor Extender (TSE) is a 6 Degree-Of-Freedom robotic mechanism for reaching high ceilings and positioning an end effector. Three scissor mechanisms are arranged in parallel, with the bottom ends coupled to linear slides, and the top vertex attached to an end effector plate. Arbitrary positions and orientations of the end effector can be achieved through the coordinated motion of the six linear actuators located at the base. By changing key geometric parameters, the TSE's design can yield a specific desired workspace volume and differential motion behavior. A general kinematic model for diverse TSEs is derived, and the kinematic properties, including workspace, singularity, and the Jacobian singular values, are evaluated. From these expressions, four key design parameters are identified, and their sensitivity upon the workspace volume and the Jacobian singular values is analyzed. A case study in autonomous aircraft assembly is presented using the insights gained from the design parameter studies. 
\end{abstract}



\section{Introduction}
\IEEEPARstart{A}{utomated} manufacturing of large products, such as aircraft and ships, is challenging for several reasons. Traditional industrial robots do not suit these factory environments, since the products are too large to transfer along a conveyor line. Robots must access manufacturing sites within large bodies, adaptively position workpieces and end-effectors relative to the large structure, and manipulate them in a complex 3-dimensional (3D) space. 

Aircraft assembly, for example, involves complex tasks where robots must access various locations across an aircraft, enter the fuselage, carry a heavy end-effector, lift workpieces to the ceiling, and mate them with 3D structures. The robot must be compact and low-profile so that it can go through narrow access routes or move across low-ceiling spaces. The robot is often required to reach a high wall or ceiling, lift objects up to these locations, and position them at these specified locations. 


This paper presents the geometric design analysis for a novel robot that is capable of lifting an object and mating it with a 3D structure. The robot structure, consisting of three pairs of scissor mechanisms and six prismatic actuators, is extendable to allow it to reach locations high above the ground, and can maneuver its end-effector to an arbitrary position and orientation, yet is low-profile once contracted. The basic mechanism of this Triple Scissor Extender is a novel integration of the Gough-Stewart hexapod platform \cite{Stewart1965} \cite{Gough1956} and the scissor lift mechanism. The former allows for 6 Degree-of-Freedom (DOF) motion with a parallel link mechanism, while the latter provides a large workspace with the highly extendable mechanism.

Parallel manipulators have been studied extensively since the seminal work by Kumar \cite{Kumar1992} and have recently been compiled in textbooks \cite{Huang2013}, \cite{Angeles2014}. The drawback of parallel manipulators is a relatively small workspace, as analyzed in \cite{Jakobovic2002}, \cite{Merlet1997}, and \cite{Wang2010}. Combining parallel manipulators with extendable mechanisms, such as cables \cite{Orekhov2015} and scissor mechanisms \cite{Chablat2016} \cite{Yang2015}, makes the workspace significantly larger.
\begin{figure}[t]
	\centering
	\includegraphics[scale=0.3]{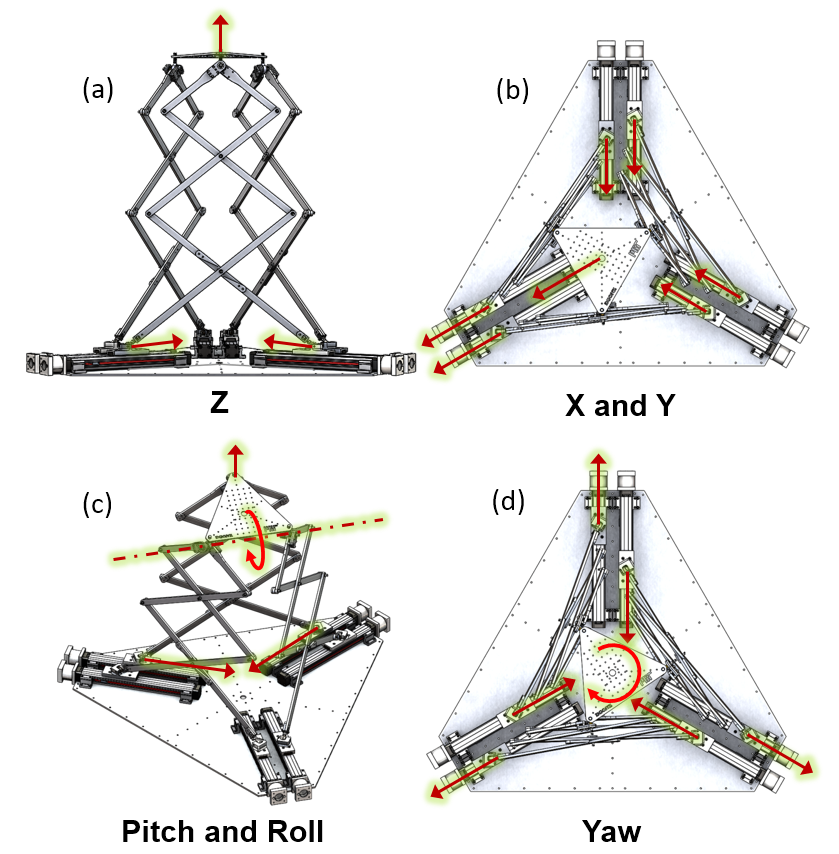}
	\caption{6-DOF motions of the Triple Scissor Extender}
	\label{MotionModes}
\end{figure}

The Triple Scissor Extender (TSE) differs from those designs in the number of DOFs and basic kinematic structure. The design concept and kinematic analysis of a particular TSE geometry has been presented in our prior work \cite{Gonzalez2016}. Here we a) present a general class of  mechanisms and their kinematic model, b) analyze kinematic performance and singularity, c) introduce design parameters and study parameter sensitivities, and d) design a prototype with an application in aircraft assembly automation.


\section{Triple Scissor Extender Mechanism}\label{MechanismDesc}
\begin{figure}[ht]
	\centering
	\includegraphics[scale=0.4]{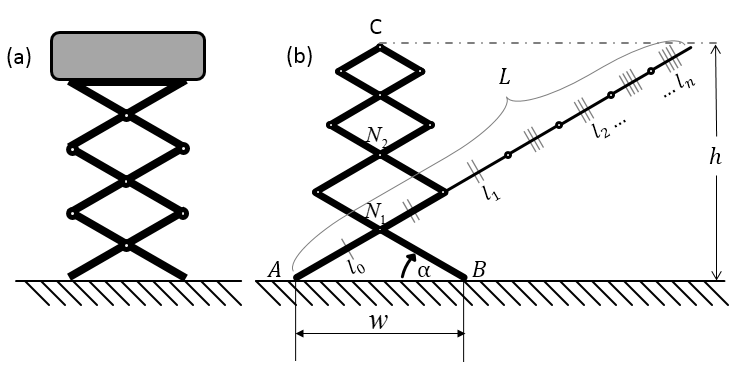}
	\caption{Kinematic parameters of a single scissor mechanism.}
	\label{SingleScissorKinematics}
\end{figure}

Scissor lifts have been used in many industries, especially in construction, for lifting workers and materials. As shown in Fig. \ref{SingleScissorKinematics}, the pantograph or scissor mechanism allows for reaching a ceiling or high position as it is extended. Once contracted, it becomes short and convenient for transportation. 

The novelty of the Triple Scissor Extender (TSE) design is to combine three sets of planar scissor mechanisms to create 6-DOF motion at the endpoint. First the tip links of the scissor mechanism are connected at point C in Fig. \ref{SingleScissorKinematics}-b, and the two bottom ends, points A and B, are moved with two independent actuators. Fig. \ref{MotionModes} illustrates a TSE with a particular layout of six linear actuators fixed to the base plate. The tip of each scissor mechanism, point C, is connected to each of the three apices of the triangular top plate through a free ball joint. Likewise the bottom ends of each scissor mechanism, points A and B, are connected to two linear slides of actuators with free ball joints.

The top plate, which is the output of the system, moves up and down, side to side, and rotates about three axes, as the six linear actuators move the six bottom ends of the three scissors, which are the inputs of the system. Fig. \ref{MotionModes}-a shows that the top plate moves upward as all six actuators move inwards, while Fig. \ref{MotionModes}-b shows that the top plate shifts sideways as the bottom ends of two scissor pairs move in opposite directions, one inward and the other outward, while the bottom ends of the third scissor move inwards. Fig. \ref{MotionModes}-c shows that the top plate makes roll and pitch rotations as one scissor moves inward, while the other two pairs of scissors are fixed. Finally, Fig. \ref{MotionModes}-d shows that a yaw motion is created with the same opposite movements of both bottom ends at the three scissors.
\begin{figure}[ht]
	\centering
	\includegraphics[scale=0.25]{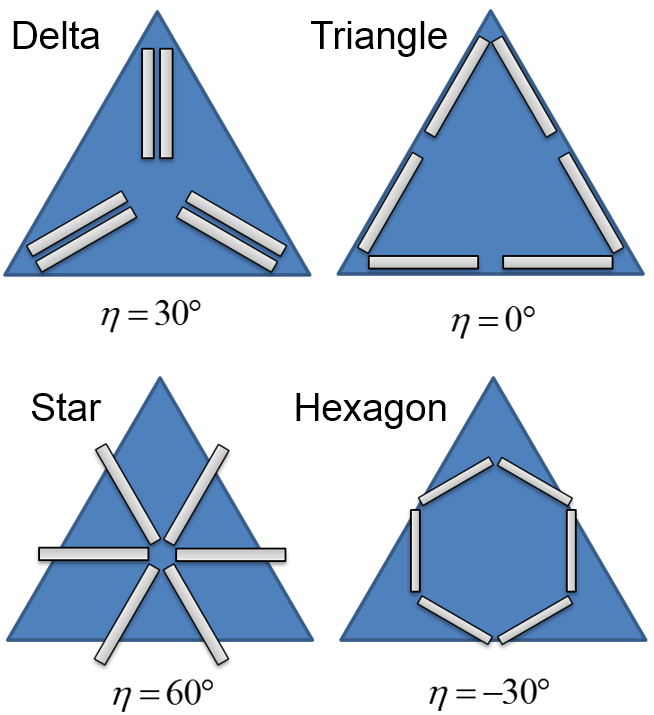}
	\caption{Various patterns of linear actuator configurations}
	\label{VariousEta}
\end{figure}  

\begin{figure}[ht]
	\centering
	\includegraphics[scale=0.20]{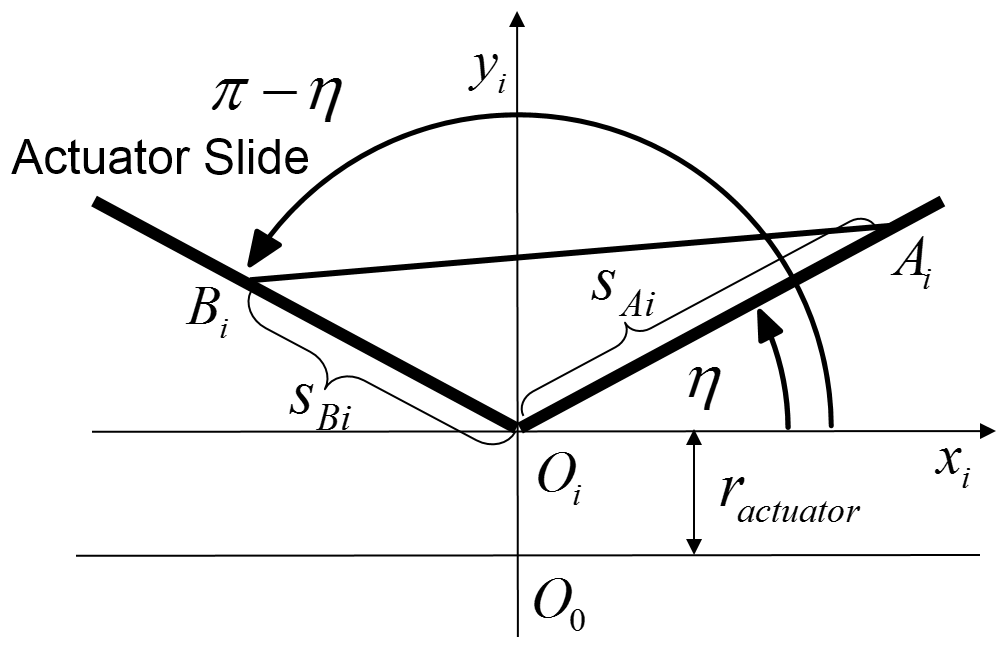}
	\caption{Top view of the scissor coordinate system and explanation of the design parameters $\eta$ and $r_{actuator}$}
	\label{EtaExplanation}
\end{figure}

These 6-DOF movements of the top plate vary depending on the layout of the six linear actuators on the base plate and other design parameters of the mechanism. Fig. \ref{VariousEta} illustrates various patterns of the linear actuator configurations. Assuming symmetry of the mechanism, it is natural to arrange each pair of the linear actuators in the same symmetric configuration, as shown in Fig. \ref{EtaExplanation}. Then the various configurations of Fig. \ref{VariousEta} can be modeled with two parameters: the orientation of the linear actuator slides and the distance between the intersection of the linear actuator slides and the center point of the entire TSE.

\begin{figure}[ht]
	\centering
	\includegraphics[scale=0.3]{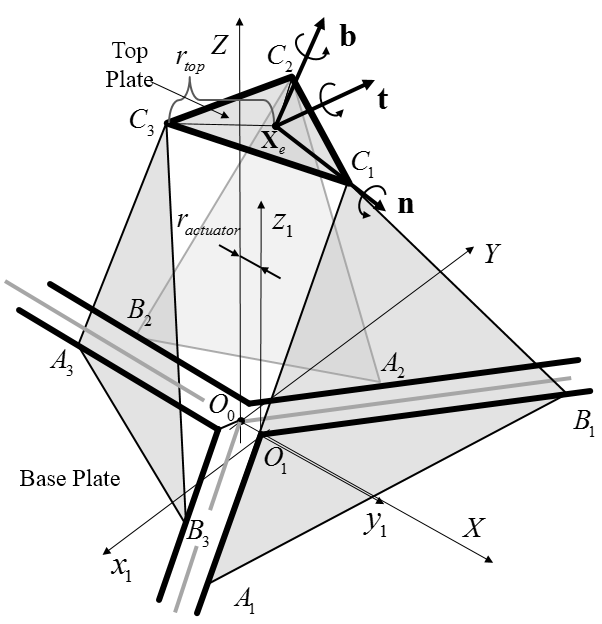}
	\caption{Base and End Effector Coordinate Systems}
	\label{TSECoords1}
\end{figure}

Fig. \ref{TSECoords1} shows the coordinate systems and the design parameters as addressed above. A world coordinate system, $O_0-XYZ$ is attached to the center point of the base plate. The three scissor mechanisms are numbered 1 to 3, and the bottom ends of the $i$-th scissor are indicated as points $A_i$ and $B_i$. The three apices of the top plate are connected to the tips of the three scissors at points $C_1$, $C_2$ and $C_3$. A local coordinate system $O_i-x_iy_iz_i$ is attached to each scissor mechanism, having its center at the intersection of the two linear actuator slides. The first scissor mechanism, for example, is described in the local coordinate system, as shown in Fig. \ref{EtaExplanation}. The configuration of the two linear actuators is described with the angle $\eta$  measured from the local $x_i$ axis and the offset distance  $r_{actuator}$ between  $O_0$ and $O_1$. Furthermore, the size of the top plate, $r_{top}$, is the distance between its center $X_{top}$ and the three apices $C_1$, $C_2$ and $C_3$.

The design parameters of the scissor mechanisms are determined by the length of the first link, $\ell_0$, and the total length of the pantograph links, $L$, as analyzed in the following section. See Fig. \ref{SingleScissorKinematics}-b. In summary the kinematic properties of the TSE are dependent on the parameters of the individual scissors, $\ell_0$ , $L$, and the size of the top plate $r_{top}$, and those of the configuration of the linear actuators, angle $\eta$ and offset $r_{actuator}$. The following sections will analyze basic kinematic relations of the general 6-DOF TSE, its kinematic properties in relation to these design parameters, and discuss applications of the robot.

\section{Kinematic Analysis} \label{Kine}
In our initial publication, we derived the kinematic model for a Triple Scissor Extender (TSE) with specific design parameters that were implemented in the prototype. Here we expand on this derivation to the general case, where the linear actuators are arranged in various patterns.  

The position of the top plate is represented with vector $X_{\epsilon}^T=\begin{pmatrix} X_{\epsilon}& Y_{\epsilon}& Z_{\epsilon}\end{pmatrix}^T$ with reference to the base coordinate system $O_0-XYZ$. The orientation of the top plate is described with roll, pitch, and yaw angles $\Theta_{\epsilon}=\begin{pmatrix}\varphi_{\epsilon} & \theta_{\epsilon} & \psi_{\epsilon} \end{pmatrix}^T$. Collectively the position and orientation of the top plate, or pose, is described as 
\begin{equation}
	p=\begin{pmatrix}X_{\epsilon}^T & \Theta_{\epsilon}^T\end{pmatrix}^T=
	\begin{pmatrix}X_{\epsilon}& Y_{\epsilon}& Z_{\epsilon}& \varphi_{\epsilon} & \theta_{\epsilon} & \psi_{\epsilon}\end{pmatrix}^T
\end{equation}

Let $s_{Ai}$ and $s_{Bi}$ be displacements of the linear actuators that move points $A_i$ and $B_i$ of the $i$-th scissor mechanism. Collectively, the six actuator displacements are described as:
\begin{equation}
	q=\begin{pmatrix}s_{A1} & s_{B1} & s_{A2} & s_{B2} & s_{A3} & s_{B3}\end{pmatrix}^T
\end{equation}

The objective of this section is to find the kinematic relationship between $p$ and $q$. For this class of complex parallel robot mechanism, however, the forward kinematic equation relating $p$ to $q$, $p = f (q)$, is prohibitively complex. Instead we aim to investigate the inverse kinematics problem. Traditional 6-DOF platforms have relatively simple inverse kinematics solutions \cite{Angeles2014}. Though the explicit, or closed-form, inverse kinematic equations, $q = f^{-1}(p)$,  are not available for the TSE, key kinematic conditions relating $p$ to $q$ can be obtained explicitly and the Inverse Kinematics can be solved numerically. Kinematic properties, such as the workspace, singularities, and the Jacobian singular values, can be evaluated directly from those kinematic conditions. 

\subsection{Inverse Kinematics}
The Inverse Kinematics (IK) problem can be worked out in the following procedure:
\begin{itemize}
	\item 
	Given the 6-dimensional pose $p$ of the top plate, obtain coordinates of points $C_1$, $C_2$, and $C_3$ in the world coordinate system $O_0-XYZ$. As shown in Fig. 5, let $\hat{n}$, $\hat{t}$, $\hat{b}$ be, respectively, the unit vectors pointing in the three directions of a Cartesian coordinate frame attached to the top plate. These unit vectors can be obtained from roll, pitch, and yaw angles $\Theta_{\epsilon}=\begin{pmatrix}\varphi_{\epsilon} & \theta_{\epsilon} & \psi_{\epsilon} \end{pmatrix}^T$.   Using these unit vectors the coordinates of the three apices are given by
	\begin{equation}\label{TopPointsWRTO}
		\begin{split}
			\vec{X}_{C1}&={}\vec{X}_{\epsilon}+r_{top}\hat{n}\\
			\vec{X}_{C2}&={}\vec{X}_{\epsilon}+r_{top}\left(\frac{-1}{2}\hat{n} + \frac{\sqrt3}{2}\hat{t}\right)\\
			\vec{X}_{C3}&={}\vec{X}_{\epsilon}+r_{top}\left(\frac{-1}{2}\hat{n} - \frac{\sqrt3}{2}\hat{t}\right)
		\end{split} 
	\end{equation}
	\item
	For each scissor mechanism $i$, given apex coordinates $C_i$, solve for the actuator displacements $(s_{Ai},s_{Bi})$. Explicit functions directly relating the actuator displacements to the apex coordinates are difficult to obtain. Implicit functional relationships, or Kinematic Constraint Equations, can be derived, as detailed in the following section. Kinematic properties can be obtained from the implicit functions and $q$ can be evaluated numerically.
\end{itemize}

\subsection{General Kinematic Constraint Derivation}
The basic 2D kinematic relationship of a single scissor mechanism can be obtained. As shown in Fig. \ref{SingleScissorKinematics} there is a functional relationship between the width of the scissors base, $w_i=\overline{A_iB_i}$, and the height of the scissors $h_i$, $i=1,2,3$. For brevity, the subscript $i$ will be omitted hereafter.
\begin{figure}[ht]
	\centering
	\includegraphics[scale=0.35]{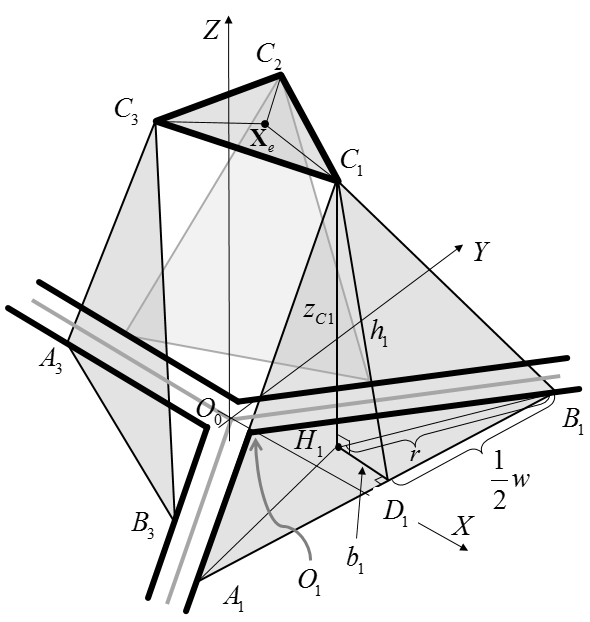}
	\caption{Projection of Relevant Points onto the Base Plane}
	\label{TSECoords2}
\end{figure}

The scissor mechanism consists of $n$ parallelograms of side length $\ell_1, \ell_2,\dots,\ell_n$, and one isosceles triangle of equal side $\ell_0$, connected at the center nodes. Let $\alpha$ be the angle of each scissor link relative to the baseline, $\alpha=\angle ABN_1$, where $N_1$ is the first connecting joint node, as shown in Fig. \ref{SingleScissorKinematics}. The width $w$ is given by
\begin{equation}\label{widthEqn}
	w=2\ell_0\cos(\alpha)
\end{equation}
Since all the links are kept parallel, the height $h$ is given by
\begin{equation}\label{heightEqn}
	h=L\sin(\alpha)
\end{equation}
where the total length $L$ is given by
\begin{equation}
	L=\ell_0+2\ell_1+\dots+2\ell_{n-1}+2\ell_n
\end{equation}
See Fig. \ref{SingleScissorKinematics} for geometric interpretation. Eliminating angle $\alpha$ from \eqref{widthEqn} and \eqref{heightEqn} yields 
\begin{equation}\label{EqA}
	\left(\frac{h}{L}\right)^2+\left(\frac{w}{2\ell_0}\right)^2=1
\end{equation}

The displacements of linear actuators, $s_{A}$ and $s_{B}$, determine the $(x,y)$ coordinates of scissor base points $A$ and $B$. From Fig. 4, we obtain:
\begin{equation}
	\begin{split}
		x_{A}=s_{A}\cos(\eta)\\
		y_{A}=s_{A}\sin(\eta)\\
		x_{B}=s_{B}\cos(\pi-\eta)\\
		y_{B}=s_{B}\sin(\pi-\eta)
	\end{split}
\end{equation}
where $\eta=\pi/6$ corresponds to the Delta configuration, $\eta=0$ for the Triangle configuration, $\eta=\pi/3$ for the $60^\circ$ Star configuration, and $\eta=-\pi/6$ for the Hexagonal configuration (See Fig. \ref{VariousEta}). Using these coordinates, the width of the scissors, $w$, can be written as
\begin{equation}
	w^2=(x_{Ai}-x_{Bi})^2+(y_{Ai}-y_{Bi})^2
\end{equation}
Note that the scissor mechanism is symmetric with respect to its centerline. Therefore, point $H_{i}$, that is, the projection of point $C$ onto the $O_i-xy$ plane, is on the bisector of the baseline $AB$. Hence, $\overline{AH}=\overline{BH}=r$, or
\begin{equation}\label{ConEq1}
	\begin{split}
		r^2&=(x_A-x_C)^2+(y_A-y_C)^2\\
		&=(x_B-x_C)^2+(y_B-y_C)^2
	\end{split}
\end{equation}
where $x_C$ and $y_C$ are the $xy$ coordinates of point $C$, i.e. those of point $H$.
Eq.(\ref{ConEq1}) is the first Kinematic Constraint Equation. 

The $z$ coordinate of point $C$ provides another condition by relating height of the scissor $h_i$ to $z_{Ci}$. Consider the right triangle $C_1H_1D_1$ in Fig. \ref{TSECoords2}, where D is center point between $s_A$ and $s_B$. We obtain
\begin{equation}\label{EqB}
	h^2=b^2+z_C^2
\end{equation}
where $b=\overline{HD}$.

By projecting the triangle on $O_i-xy$ plane as in Fig. \ref{TSECoords2} we obtain
\begin{equation}\label{EqC}
	r^2=b^2+\left(\frac{w}{2}\right)^2
\end{equation}
Eliminating $h$, $r$, and $b$ from \eqref{EqA}, \eqref{EqB}, and \eqref{EqC} yields the following second Kinematic Constraint Equation. 
\begin{equation}\label{ConEq2}
	\begin{split}
		(x_A-x_C)^2+(y_A-y_C)^2+z_C^2=\\
		L^2+\frac{1}{4}\left(1-\left(\frac{L}{\ell_0}\right)^2\right)\left((x_{A}-x_{B})^2+(y_{A}-y_{B})^2\right)
	\end{split}
\end{equation}

Simultaneous equations \eqref{ConEq1} and \eqref{ConEq2} determine actuator displacements $(s_A, s_B)$ for given coordinates $(x_C, y_C, z_C)$. Repeating the same computation for the three pairs of scissors mechanisms determines the six actuator displacements for a given top plate position and orientation.

\subsection{Singularity Analysis}\label{SingularityAn}


In general the proposed TSE can move the top plate in an arbitrary direction in the 6 dimensional space. However, singularity limits its "dexterity" at certain configurations. Here we obtain conditions for type-1 singularities \cite{56660}, when input motion leads to zero output motion, when the Jacobian relating the top plate velocity $\dot{p}$ to actuator velocities $\dot{q}$ exists: $\dot{p}=J\dot{q}$. In this study, we do not investigate type-2 singularities, when output motion may be achieved with zero input motion. Singularity of the TSE occurs at particular combinations of a) actuator displacements, b) top plate position and orientation, and c) structural parameters of the mechanism, including $\ell_0$ and $\eta$, where nonzero actuator velocities $\dot{q}\neq0$ Produce no velocity at the top plate, $\dot{p}=0$. Since the top plate can move in all 6 dimensional directions only when each of the three apices can exhibit 2 DOF motion with the two actuators, it can be shown that the necessary and sufficient conditions for singularity of TSE are given by
\begin{equation}
	\exists \quad \dot{s}_{Ai} \quad \& \quad \dot{s}_{Ai} , \quad i\in\left( 1,2,3 \right)
\end{equation}
such that
\begin{equation}
	\dot{x}_{Ci} = \dot{y}_{Ci} = \dot{z}_{Ci} = 0, \quad \& \quad \dot{s}_{Ai} \dot{s}_{Bi} \neq 0
\end{equation}
This implies that the TSE singularity can be analyzed with the individual scissor mechanisms governed by the two constraint equations (\ref{ConEq1}) and (\ref{ConEq2}). Namely, i) differentiating (\ref{ConEq1}) and (\ref{ConEq2}) with respect to time, ii) setting the apex velocity to zero, $\dot{x}_{Ci}=\dot{y}_{Ci}=\dot{z}_{Ci}=0$, and iii) collecting terms, we can obtain the following conditions that actuator velocities $\dot{s}_{Ai}$ and $\dot{s}_{Bi}$ do not produce any velocity at apex $C_i$, where $i$ is 1, 2 or 3. 
\begin{equation}
	\begin{pmatrix} a_{11} & a_{12} \\ a_{21} & a_{22} \end{pmatrix}\cdot
	\begin{pmatrix} \dot{s}_{Ai} \\ \dot{s}_{Bi} \end{pmatrix}=
	\begin{pmatrix} 0 \\ 0 \end{pmatrix}
\end{equation}
where
\begin{equation}
	\begin{split}
		a_{11}&=-s_{Ai}+x_{Ci}\cos{\eta}+y_{Ci}\sin{\eta}\\
		a_{12}&=s_{Bi}+x_{Ci}\cos{\eta}-y_{Ci}\sin{\eta}\\
		a_{21}&=a_{11}+\left(1-\frac{1}{4}\left(\frac{L}{\ell_0}\right)^2\right)\left(s_{Ai}+s_{Bi}\cos{2\eta}\right)\\
		a_{22}&=\left(1-\frac{1}{4}\left(\frac{L}{\ell_0}\right)^2\right)\left(s_{Bi}+s_{Ai}\cos{2\eta}\right)\end{split}
\end{equation}
For the above linear simultaneous equation to possess a nonzero solution $\begin{pmatrix}\dot{s}_{Ai} & \dot{s}_{Bi}\end{pmatrix}^T\neq0$ the matrix must be singular, that is $a_{11}a_{22}-a_{12}a_{21}=0$. Solving this for the structural parameter $L/\ell_0$ yields

\begin{equation}\label{SingCond}
	\frac{L^2}{\ell_{0}^2}=1+\frac{4a_{11}a_{12}}{a_{12}\left(s_{Ai}+s_{Bi}\cos{2\eta}\right)-a_{11}\left(s_{Bi}+s_{Ai}\cos{2\eta}\right)}
\end{equation}

\begin{figure}[t]
	\centering
	\includegraphics[scale=.375]{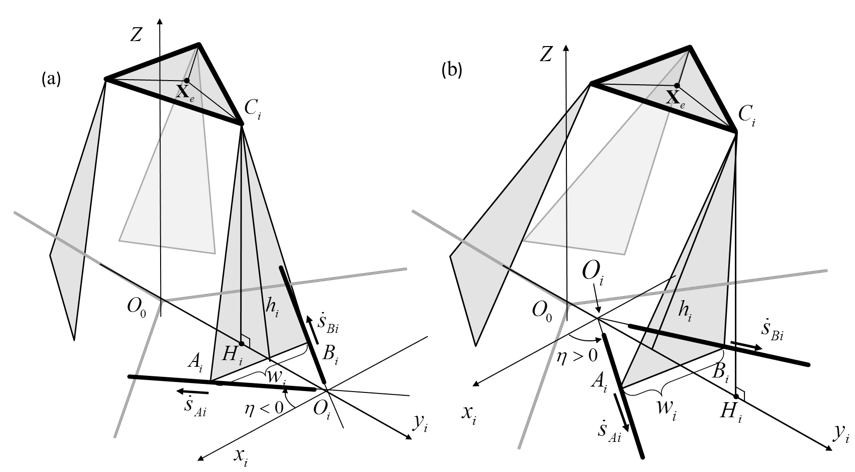}
	\caption{Explanation of Internal Singularities of the Triple Scissor Extender for (a) $\eta<0$ and (b) $\eta>0$.}
	\label{SingExplanation}
\end{figure}

Fig.\ref{SingExplanation} illustrates two singular configurations. When the actuator angle is negative, $\eta<0$, as in the case of the Hexagon layout in Fig. \ref{VariousEta}, inward actuator movements reduce the scissor width $w_i$ at the base, which reduces the height of the scissor depending on the ratio $\ell_0/L$. This downward motion, however, can be canceled out by the upward motion of the apex since the plane of the scissor tends to rise as the actuators move inwards, which tends to move the apex upwards. As the two effects cancel out, no velocity is observed at the apex, although the two actuator velocities are nonzero. For a positive actuator angle,  $\eta>0$, this type of cancellation occurs only at the top plate location away from the center line (axis Z), as shown in Fig.\ref{SingExplanation}-b. Such a singular point may be outside the stroke of the actuator. The singular configuration can also be avoided by selecting the ratio $\ell_0/L$ appropriately.

\section{Design Parameter Studies}\label{DPS}
\subsection{Computation of Kinematic Properties}\label{Computation}
\begin{figure}[ht]
	\centering
	\subfloat[Angled View]{
		\includegraphics[scale=.54]{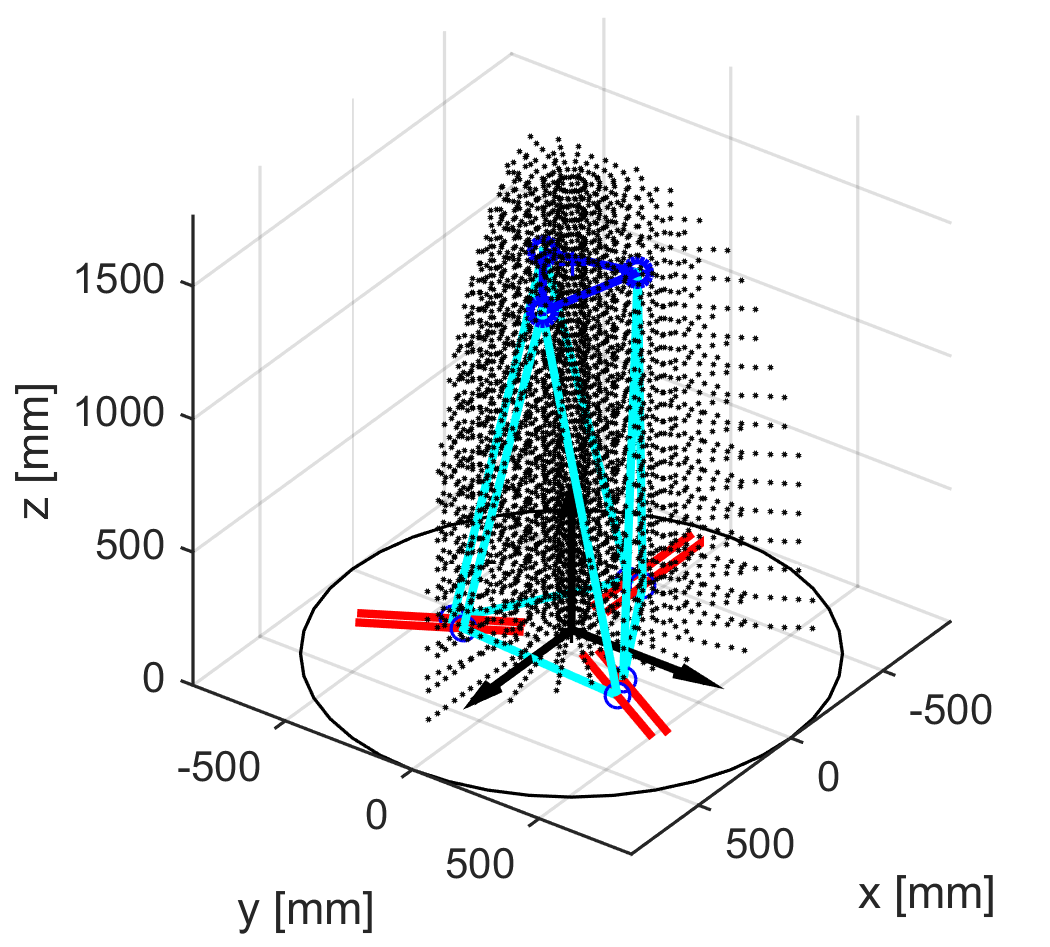}}
	\subfloat[Top View]{
		\includegraphics[scale=.4]{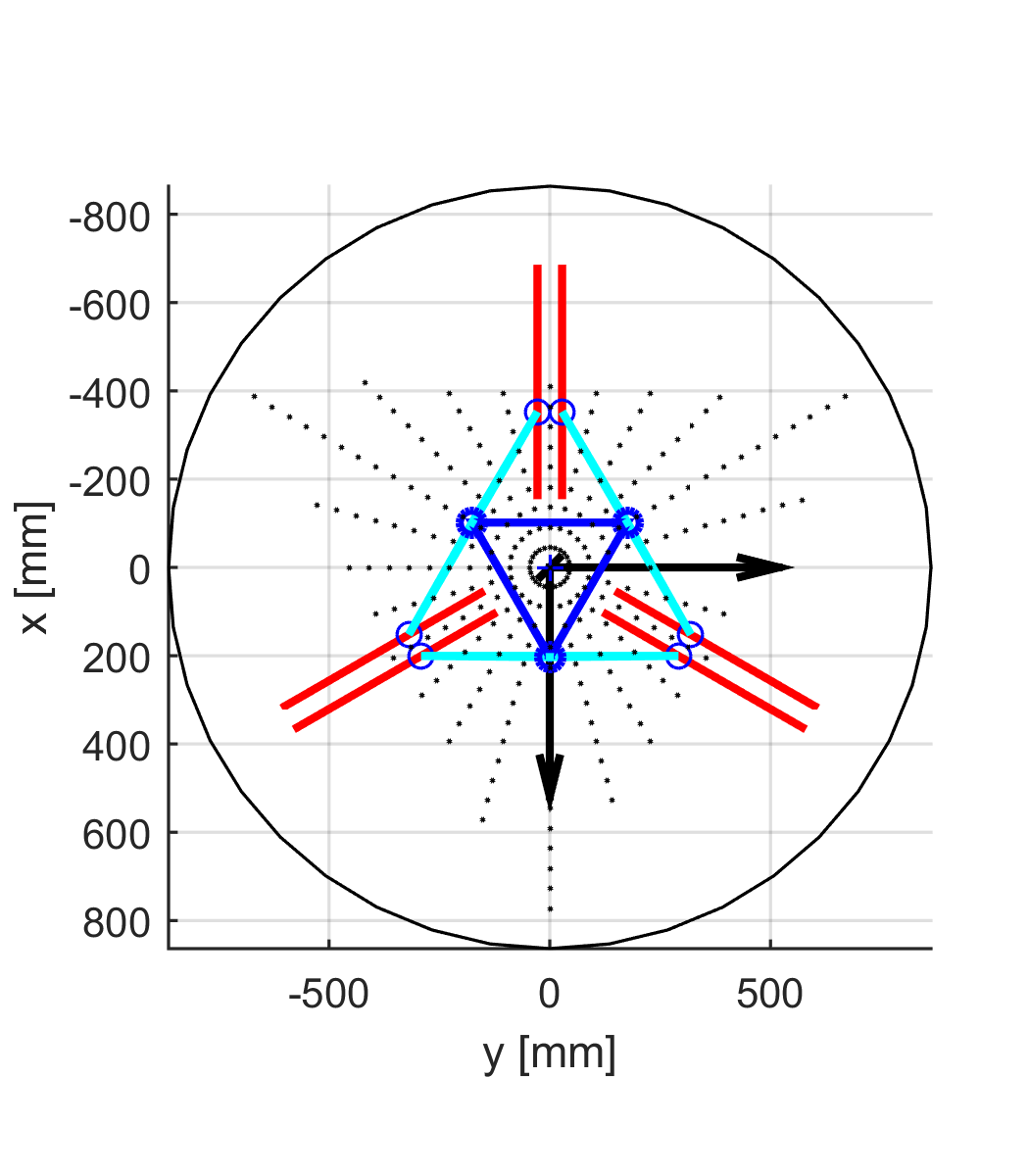}}\\
	\subfloat[Side View]{
		\includegraphics[scale=.48]{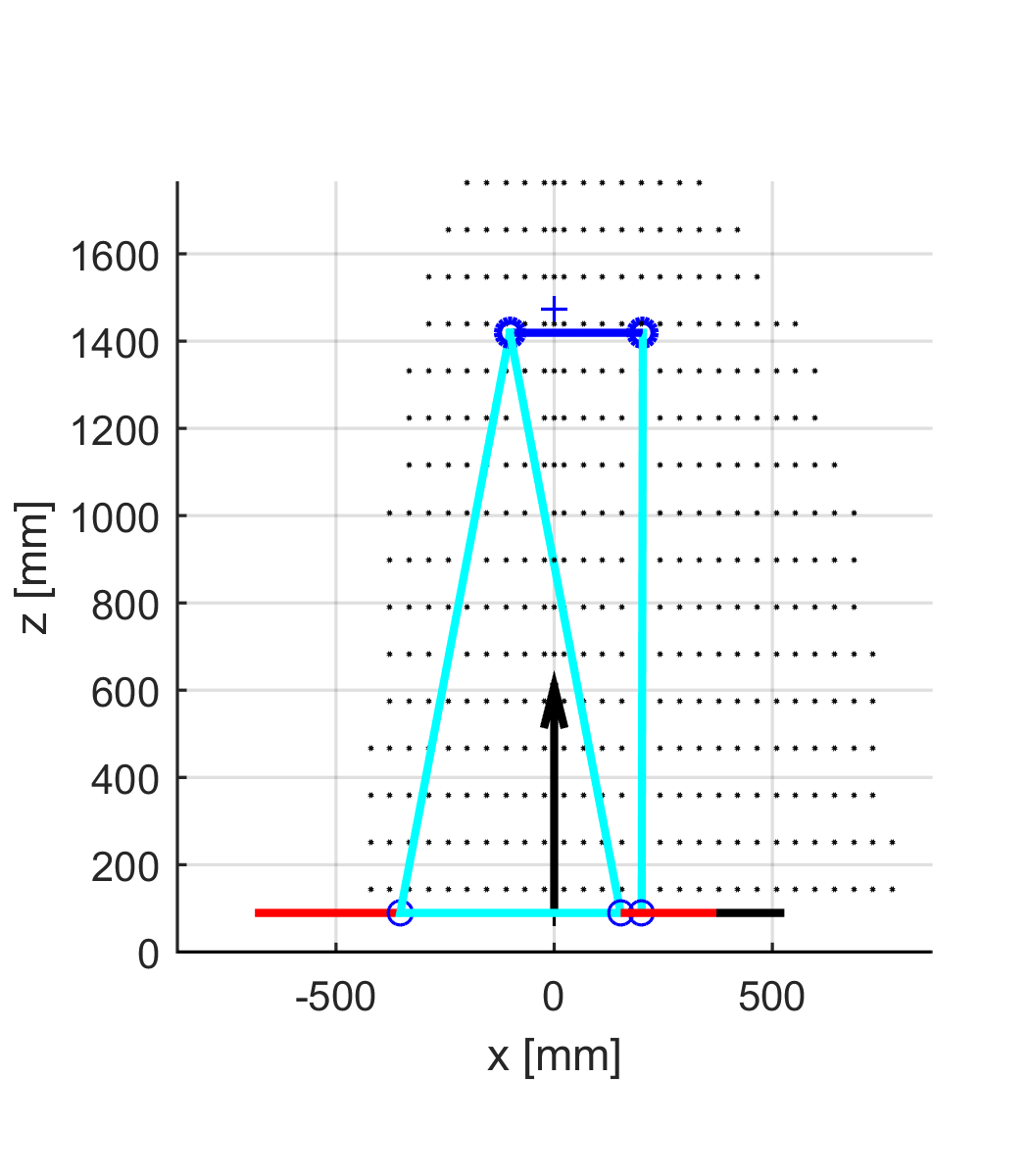}}
	\subfloat[Side View, $75^\circ$ rotation]{
		\includegraphics[scale=.48]{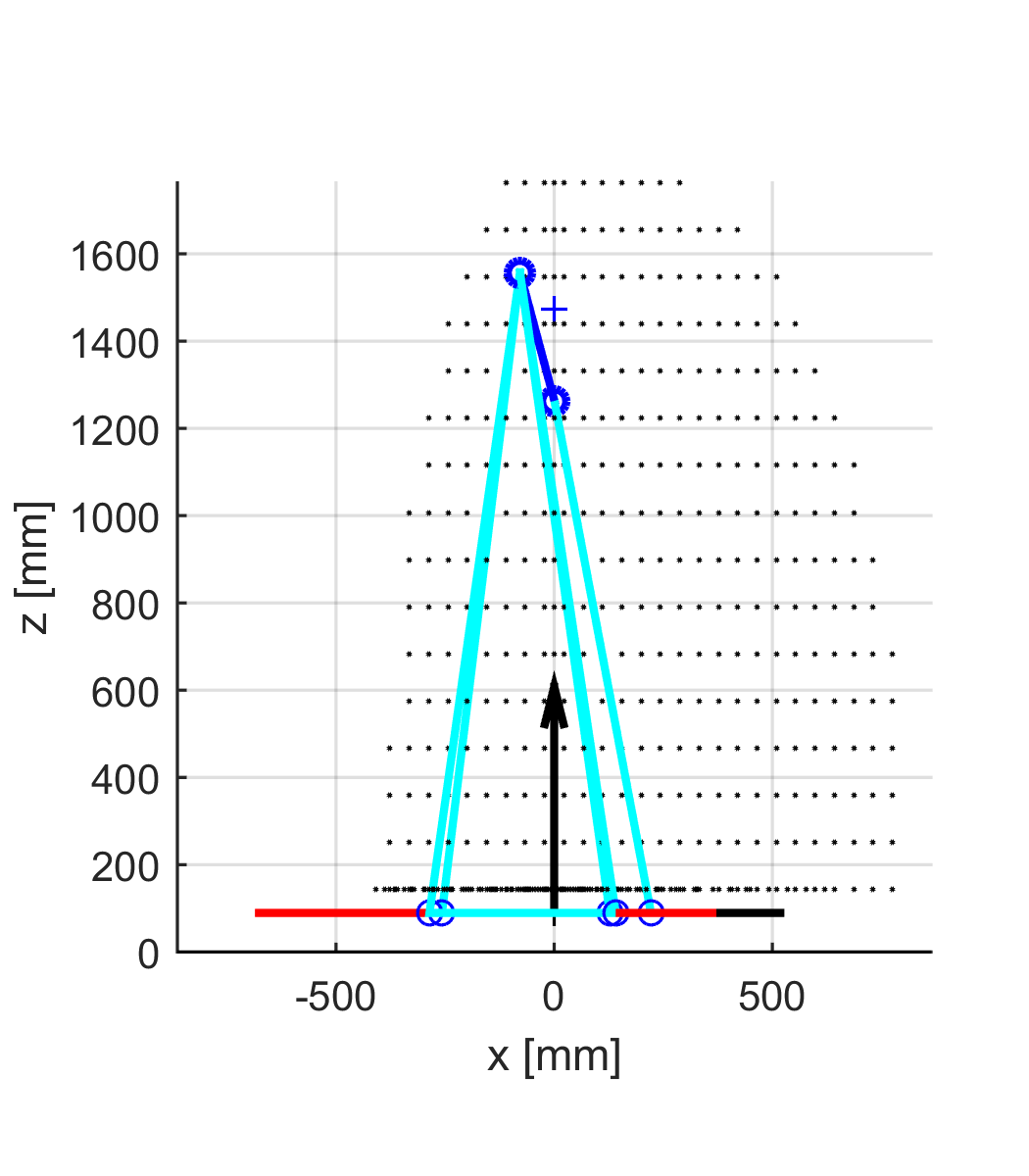}}
	\caption{Rendering of TSE Prototype Workspace.}
	\label{WorkspaceRender}
\end{figure}

Based on the Kinematic Constraint Equations obtained previously, various kinematic properties can be evaluated through numerical analysis. The reachable space of a robot endpoint, or Workspace, is a fundamental kinematic property of robotic mechanisms. The envelope of the workspace is obtained numerically through the following procedure. First, a set of test points is generated in cylindrical coordinates $\left(r,\phi,z\right)$. For our analysis, the radius $r$ ranges from $0 : L/2$, the angle $\phi$ ranges from $0 : 2\pi$, and the height $z$ ranges from $0 : L$. These points were all converted to Cartesian coordinates relative to the robot origin, and the inverse kinematics are calculated for each. From the top plate pose, the three apices are computed with \eqref{TopPointsWRTO}. The two Kinematic Constraint equations are then solved for actuator displacements $\left(s_{Ai},s_{Bi}\right)$. A test point is considered invalid if any of the actuator displacements is negative or complex as well as out of the linear actuator stroke. Such a point is not reachable and is not added to the list of valid points. The workspace envelope can be determined by repeating this examination for all the test points. Fig. \ref{WorkspaceRender}-a, b, and c illustrate the workspace computed by examining the solutions to the Kinematic Constraint equations for the horizontal top plate configuration where $\Theta_{\epsilon}=0$. Fig. \ref{WorkspaceRender}-d shows a side view of the workspace when the top plate is tilted about the Y axis, $\theta_{\epsilon}=75^\circ$.

Analysis of differential motion based on the Jacobian matrix relating the top plate velocity to the actuator velocities provides useful insights. The Jacobian can be computed for each top plate pose by differentiating the two Kinematic Constraint Equations in a way similar to the singularity analysis. Details can be found in \cite{Gonzalez2016}. Examining all the singular values of the Jacobian reveals the ability of TSE to move in 6-dimensional space. Similar analysis has been reported for different types of parallel manipulators \cite{Merlet2007}. Of particular interest for the current work is the Jacobian singular value in the vertical direction, $\sigma_z$, which can be considered the Spatial Gear Ratio between actuator motion and upward motion of the top plate. The highly extendable vertical motion is a salient feature of TSE. Numerical analysis reveals that the vertical singular value $\sigma_z$ varies widely depending on the height, while the singular values in the $XY$ horizontal directions are sufficiently large at diverse heights within a non-singular workspace. It is important to examine each TSE design in terms of these kinematic properties.

\subsection{Design Parameters and Performance Metrics}\label{Perf}

In order to perform generic analysis of the kinematic properties of the TSE, the Kinematic Constraint Equations must be distilled to the most essential design parameters. We non-dimensionalize equations \eqref{ConEq1} and \eqref{ConEq2} by scaling them by the total scissor length parameter $L$. 

This leads to two new Kinematic Constraint Equations and four design parameters that define the entire design space for this class of machine:
\begin{itemize}
	\item $k_1$: $(\ell_0/L)$, the ratio of the lowest scissor link length and the total scissor length $L$. See Fig. \ref{SingleScissorKinematics}.
	\item $k_2$: $\eta$, the angle of the actuator from the horizontal axis of its respective coordinate frame. See Figs. \ref{EtaExplanation} and \ref{VariousEta}.
	\item $k_3$: $(r_{actuator}/L)$, the ratio of the actuator coordinate radial distance from center and $L$. See Figs. \ref{EtaExplanation} and \ref{TSECoords2}.
	\item $k_4$: $(r_{top}/L)$, the ratio of the top platform distance to the ball joint and the total scissor length $L$. See Fig. \ref{TSECoords2}.
\end{itemize}

The scissor top points $x_C$, $y_C$, and $z_C$ as well as the linear slide positions $S_A$ and $S_B$ are also normalized over the total scissor length $L$, which we define as $\hat{x_C}$, $\hat{y_C}$, $\hat{z_C}$, $\hat{S_A}$, and $\hat{S_B}$. 

By rearranging and replacing variables in equations \eqref{ConEq1} and \eqref{ConEq2}, the Non-Dimensionalized Kinematic Constraint Equations emerge. For the Delta configuration of linear actuators, they are given by
\begin{equation}\label{NDKin1}
	\begin{split}
		\hat{S_A}^2 - 2\hat{x_C}\hat{S_A}\cos(k_2)+2(k_3-\hat{y_C})\hat{S_A}\sin(k_2)=\\
		\hat{S_B}^2 + 2\hat{x_C}\hat{S_B}\cos(k_2)+2(k_3-\hat{y_C})\hat{S_B}\sin(k_2)
	\end{split}
\end{equation}
and
\begin{equation}\label{NDKin2}
	\begin{split}
		&\hat{S_A}^2 +2(k_3-\hat{y_C})\hat{S_A}\sin(k_2)+k_3^2 - 2\hat{y_C}k_3+\\
		&\hat{x_C}^2+\hat{y_C}^2+\hat{z_C}^2 = 1+2\hat{x_C}\hat{S_A}\cos(k_2)+\\
		&\frac{1}{4}\left(1-\left(\frac{1}{k_1}\right)^2\right)\left(\hat{S_A}^2+\hat{S_B}^2+2\hat{S_A}\hat{S_B}\cos(2k_2)\right)
	\end{split}
\end{equation}

In order to judge one design of TSE against another, we use two performance metrics: the Height Amplification Factor $(HAF)$, and the Workspace Volume Percentage $(\%_{ws})$. 

The Workspace Volume Percentage $(\%_{ws})$ is the percentage of the number of reachable test points examined within the cylindrical volume described previously. 

The Height Amplification Factor ($HAF$) is given by
\begin{equation}
	HAF = \frac{\sigma_z(z;k_i)}{\sigma_z(z;k_i=k_{i,0})}
\end{equation}
where $k_{i,0}$ is a constant design parameter selection to which all others are compared ($k_{1,0}=1$, $k_{2,0}=0^\circ$, etc). Note that by dividing the $z$-dependent singular value by another z-dependent singular value, the $HAF$ becomes constant throughout the $z$-range, and is no longer dependent on the configuration variable $z$. The $HAF$ may then be plotted directly against the design parameter $k_i$ for comparison.

\subsection{Sensitivity Analysis}\label{DPSensitivity}
In the previous subsection the Non-dimensionalized Kinematic Constraint Equations \eqref{NDKin1} and \eqref{NDKin2} were derived in order to facilitate the analysis of the kinematic properties of the TSE. In this section, we analyze how changing the kinematic design parameters $k_1$, $k_2$, $k_3$, and $k_4$ affect the configuration-dependent Translational Spatial Gear Ratios of the TSE and the Workspace Volume Percentage as defined previously.


For these analyses, the design parameters chosen match those of the implemented prototype and a single design parameter was changed each time to determine how deviating from this design affects the HAF and Workspace Volume Percentage. 

\subsubsection{Sensitivity due to scissor length ratio $k_1$}

The scissor length ratio $k_1$ carries with it the essence of TSE: height amplification using scissor mechanisms. Parameter $k_1$ defines the smallest footprint taken up by the TSE. The smallest possible circular area that a TSE could occupy has a radius of $\ell_0$. 

\begin{figure}[ht]
	\centering
	\includegraphics[scale=.95]{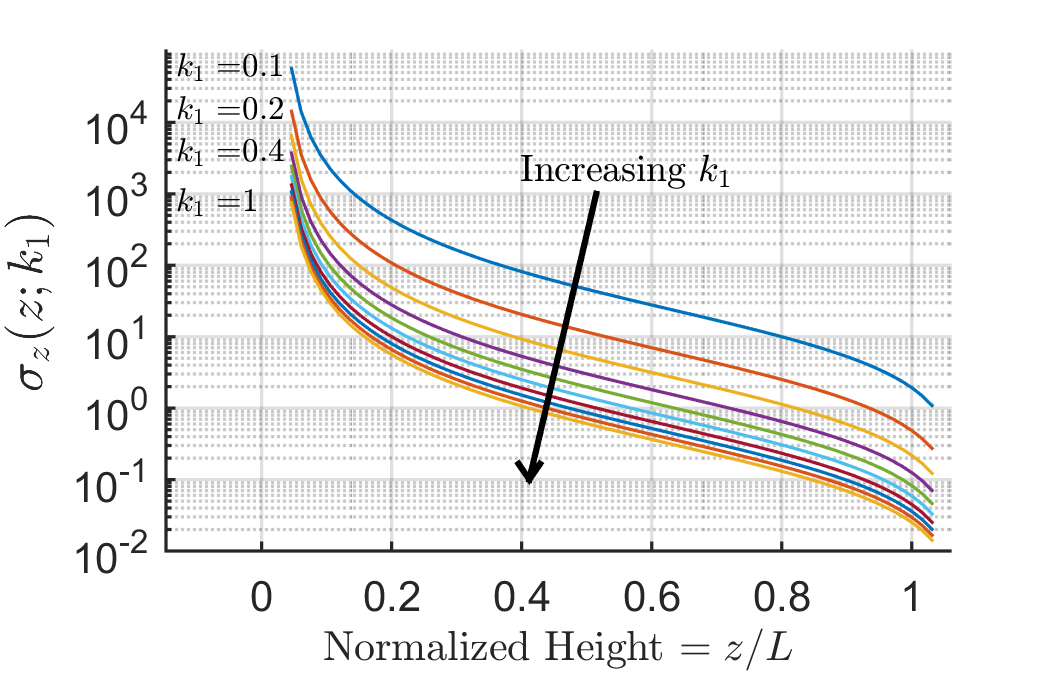}
	\caption{Jacobian Singular Values with changing Length Ratio}
	\label{ZvsK1}
\end{figure}

\begin{figure}[ht]
	\centering
	\includegraphics[scale=.8]{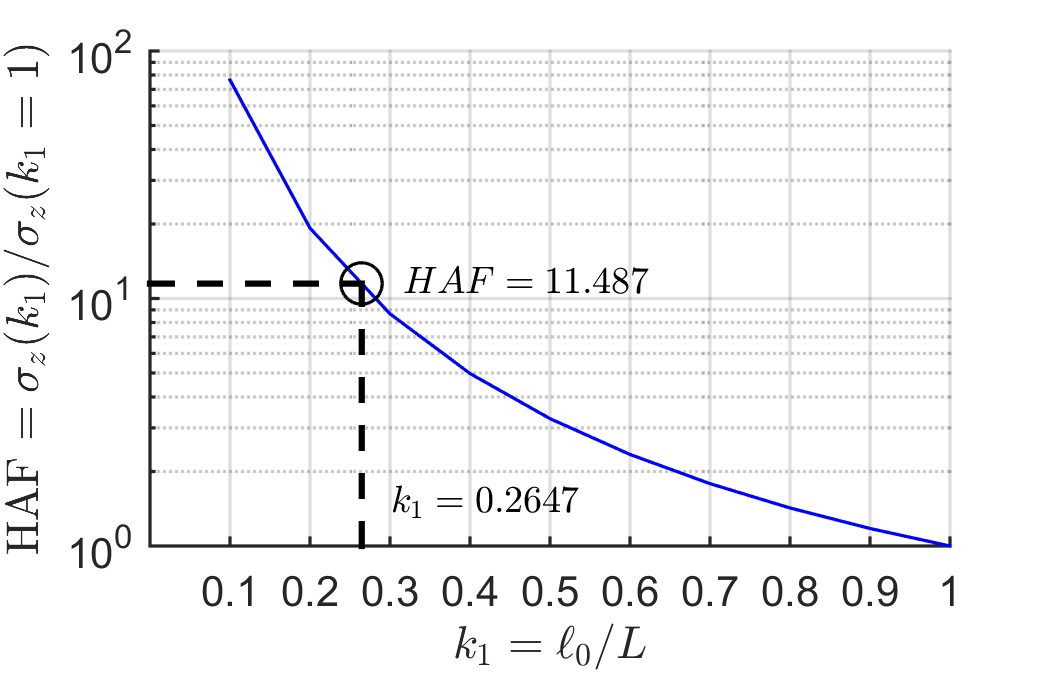}
	\caption{Height Amplification Factor with changing Length Ratio}
	\label{HAFvsK1}
\end{figure}

As can be seen in Figs. \ref{ZvsK1} and \ref{HAFvsK1}, the lower the value of $k_1$, the greater the sensitivity to motion and Height Amplification Factor. The implemented Prototype has a $k_1$ of $0.2647$, which gives a $HAF$ of $11.487$.

A $k_1=1$ has an equal total scissor length and lowest link length, $L=\ell_0$, which has no scissor mechanism at all. This configuration of the TSE is nearly identical to many other Hexapod platforms \cite{Stewart1965} \cite{Huang2013} with rigid links, and has no amplification factor ($HAF = 1$).

\begin{figure}[ht]
	\centering
	\includegraphics[scale=.95]{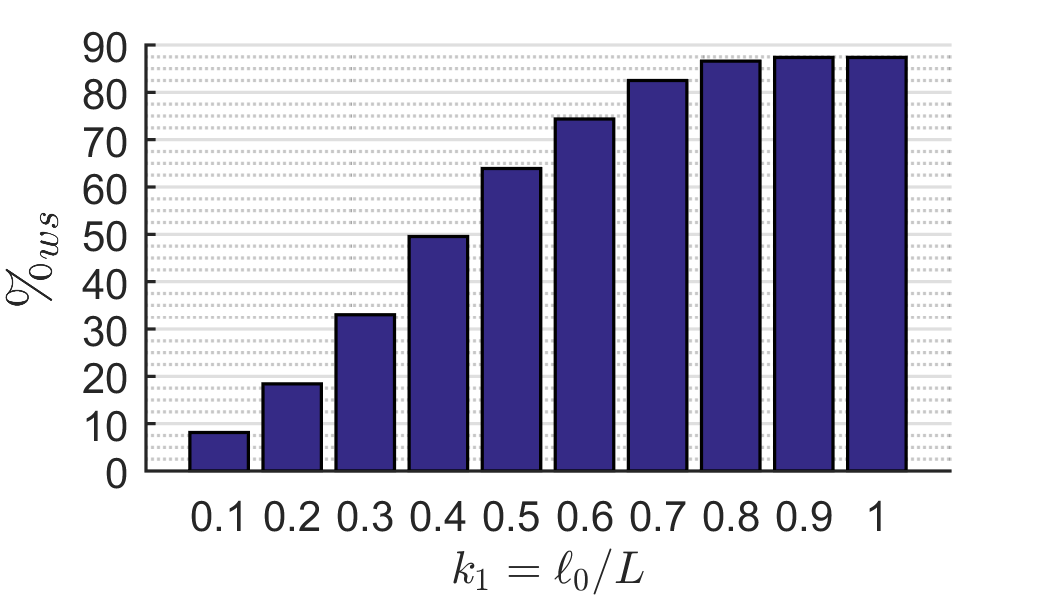}
	\caption{Workspace Volume Percentage with changing Length Ratio}
	\label{WSVsK1}
\end{figure}

The Workspace Volume decreases as $k_1$ decreases and the $HAF$ increases. Intuitively, as $\ell_0$ grows smaller with $k1$, each scissor mechanism reaches its stroke limit sooner, which in turn limits the achievable planar X and Y motion of the TSE. See Fig.\ref{WSVsK1}. Approximately, a larger Workspace Volume Percentage implies a wider workspace in the X and Y directions.

This presents a design trade-off: By decreasing $k_1$, the $HAF$ increases and the TSE can become more compact when fully retracted, but at the cost of overall Workspace Volume. If compactness is a requirement, a small $k_1$ should be used, with the lack of larger XY motion, which could be supplemented by putting the TSE on a wheeled base. 

\subsubsection{Sensitivity due to Actuator Angle $k_2$}
Intuitively, by changing the scissor angle $k_2=\eta$, we affect the rate of change of the width between the bottom points of the scissors, which directly affects the Z Jacobian Singular Values. We explored the possibility of having both positive and negative $k_2$. Angles in the interval of  $-90\degree<k_2<90\degree$ were examined for performance evaluation. 


\begin{figure}[ht]
	\centering
	\includegraphics[scale=.8]{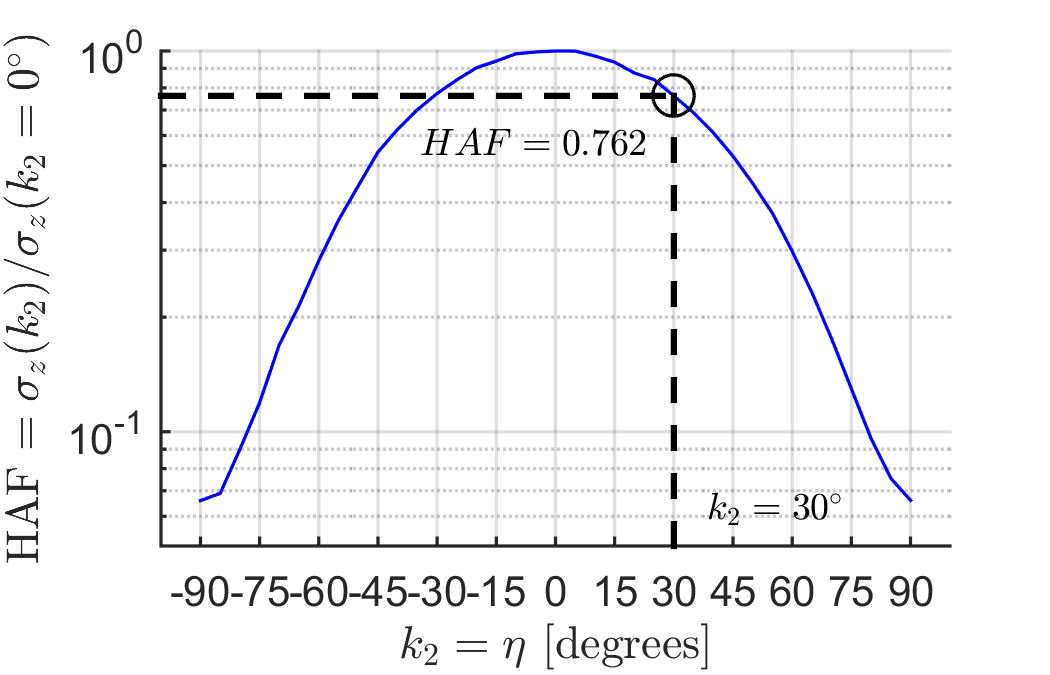}
	\caption{Height Amplification Factor with changing Slide Angle}
	\label{HAFvsK2}
\end{figure}

As shown in Fig. 
\ref{HAFvsK2}, completely aligned slides where $k_2=0^\circ$ have the largest Z Jacobian Singular Value and $HAF$, and any deviation from this will decrease this value. This is due to all motion of the linear actuators contributing to changing the width of the scissor bottom points. When the absolute value of $k_2$ approaches $90^\circ$, almost no motion of the linear actuators contributes to changing the width of the scissor bottom points, leading to poor height amplification, and a low $HAF$. Furthermore, as the absolute value of $k_2$ becomes larger and gets closer to $90^\circ$, the singularity conditions obtained previously, Eq.\ref{SingCond}, are likely to be met. With a $k_1$ of $0.2647$, TSE configurations with $k_2$ of $60^\circ$ or greater exhibit behaviors consistent with numerical instability of the kinematics solver near the singular configuration. As depicted in Fig.\ref{SingExplanation}, the upward motion of at least one scissor tends to vanish, leading to very small $\sigma_z$. As shown in Fig.\ref{SingExplanation}-a, singular configurations appear near the center line (the Z axis) for negative $k_2=\eta<0$, while they appear away from the center line for positive $k_2>0$. See Fig.\ref{SingExplanation}-b. Therefore, choosing a positive, small $k_2$ is advisable. Positive and negative values of $k_2$, while significantly changing the appearance of the TSE, exhibit nearly identical behavior in the $z$-direction. 


\begin{figure}[ht]
	\centering
	\includegraphics[scale=.95]{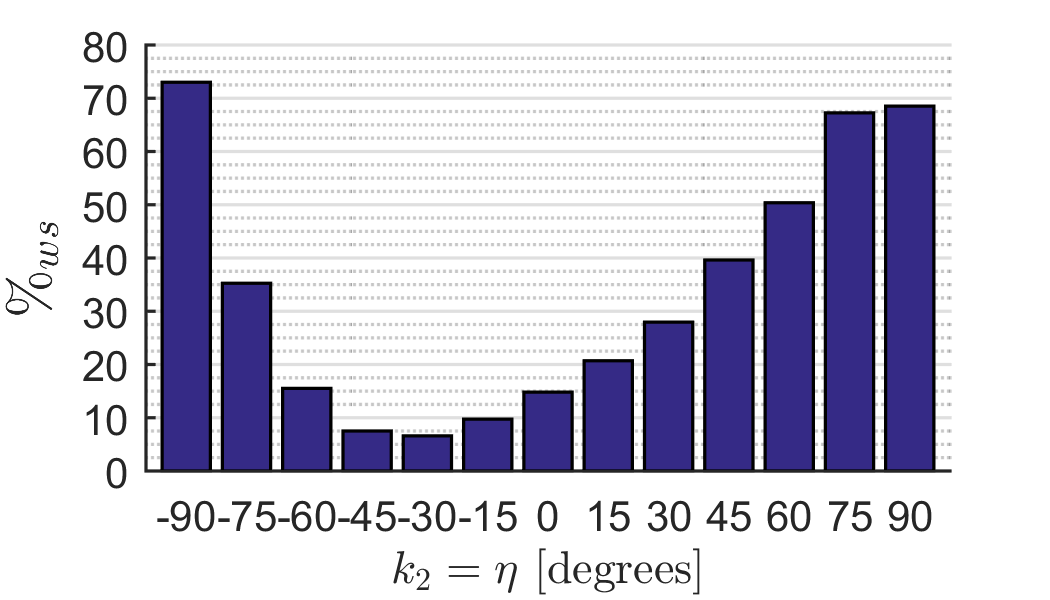}
	\caption{Workspace Volume Percentage with changing Slide Angle}
	\label{WSvsK2}
\end{figure} 

The Workspace Volume generally increases as the absolute value of $k_2$ increases (See Fig. \ref{WSvsK2}), with a small offset due to the distance of the actuator coordinate system $r_{actuator}$ from the center of the TSE. Again, a design trade-off exists for $k_2$ between the compactness offered by a large $HAF$ and the XY-planar range of motion offered by a large Workspace Volume. 

\subsubsection{Sensitivity of design parameters $k_3$ and $k_4$}\label{k3_analysis}

The numerical analysis reveals that the actuator coordinate radial distance $k_3$ does not significantly influence the Z Jacobian singular values. They scale linearly with $k_3$, but the singular value profile remains unchanged. The achievable workspace volume also does not change significantly with $k_3$. Design parameter $k_4$, which determines the top plate size, does not appear in the Kinematic Constraint Equations, but only in the first step of the Inverse Kinematic problem where the apex points are determined. Parameter $k_4$ affects the rotational spatial gear ratios; as the top plate becomes smaller, the rotational gear ratios increase in general.

Both design parameters, however, play an important role in avoiding singularity and other special configurations. As shown in Fig.\ref{SingExplanation}-b where $k_2(=\eta)$ is positive, singularity may occur when one apex of the top plate is away from the $Z$ axis. To prevent this from occurring, the radial distance of the actuators should be sufficiently larger than the size of the top plate, $k_3>k_4$. However, another type of unwanted special configurations may occur, if the top plate size $k_3$ is too small. Fig. \ref{WorkspaceRender}-d shows a configuration where the top plate (tilted $75^\circ$) is almost aligned with one of the scissor plane. If the scissor and the top plate are completely aligned, the top plate cannot be moved in the direction perpendicular to its plane at a non-zero velocity. At this configuration, the inverse Jacobian matrix is degenerate and, therefore, the forward kinematics Jacobian does not exist. The Stewart mechanism and other parallel manipulators exhibit similar behaviors. Details can be found in \cite{Huang2013}. For the TSE, another design trade-off must be considered for selecting $k_3$ and $k_4$ by considering these requirements.  


\section{Prototyping and Application} \label{CaseStudy}
With its large height range, TSE is considered for use in an aircraft manufacturing task. Assembly inside the fuselage of commercial aircraft requires a new type of robot that can reach any location along the inside of the fuselage and fine position an end-effector relative to the structure. The robot must be low profile for easy transportation and, once inside the fuselage, extend to access the walls and ceiling. 
\begin{figure}[ht]
	\centering
	\includegraphics[scale = 0.4]{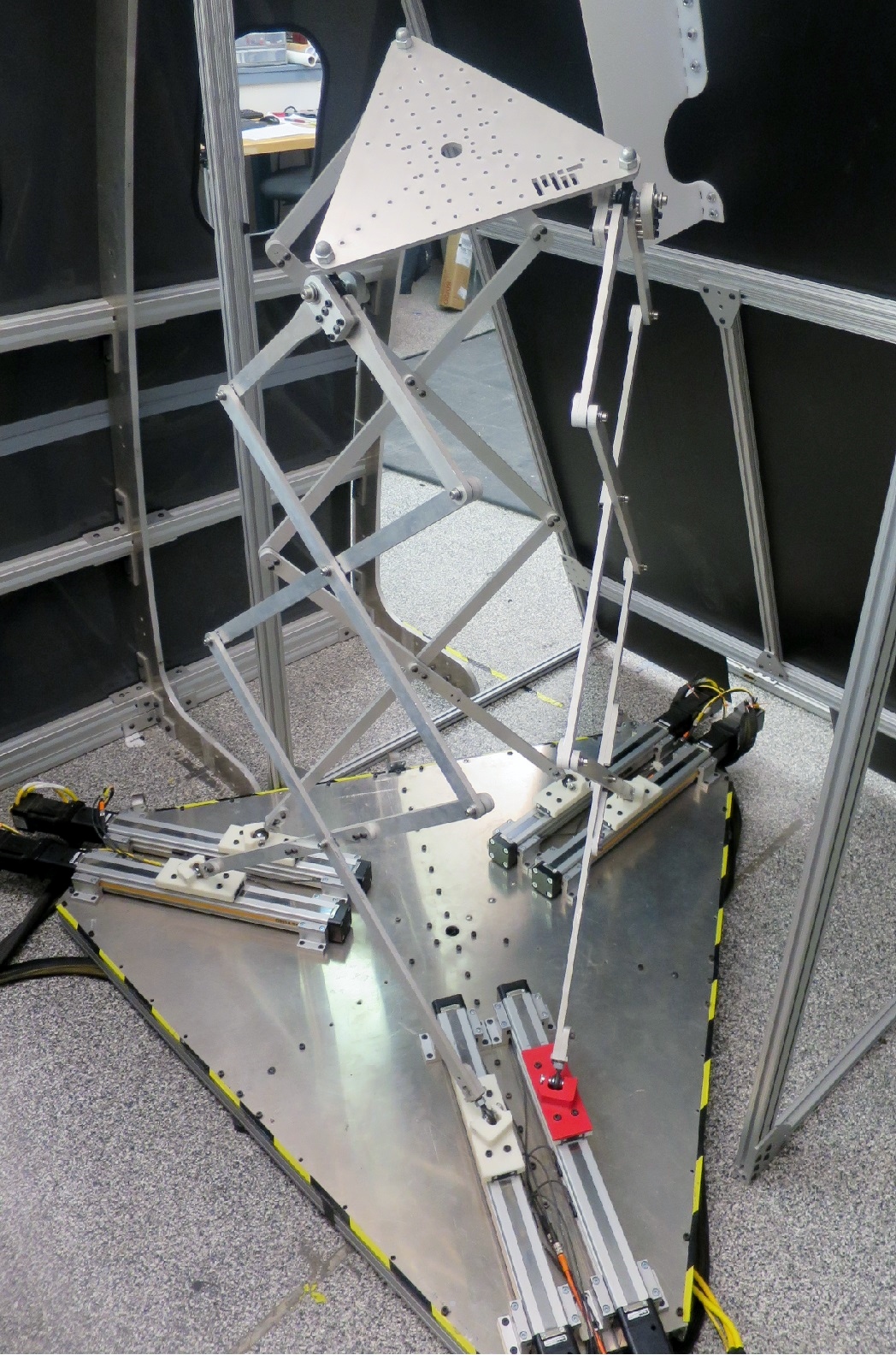}
	\caption{The Triple Scissor Extender prototype.}
	\label{TSEProto}
\end{figure}
The Triple Scissor Extender prototype (See Fig. \ref{TSEProto}) was designed to have a maximum height of 1.619 meters, roughly that of a human, and a fully retracted height of 0.324 meters for easy transportation in a laboratory setting. The minimum and maximum scissor angle $\alpha$ are limited with a mechanical stop in order to avoid the upper and lower singular configurations. This leads to a total scissor length $L$ of 1.727 meters. In addition to the height range specification, the TSE must be able to fine-position the end-effector by moving in the X-Y plane. With a workspace volume percentage of 0.28, the prototype TSE can move 1.2~1.5 m in the X and Y directions.

In order to keep the footprint small, $k_1=\ell_0/L$ was chosen to keep the total width of the base roughly half of the total scissor length $L$. Actuator slide angle $k_2=\eta$ was chosen to avoid singularity as well as to retain the necessary workspace. Top Plate Radius $k_4 = r_{top}/L$ was chosen such that it may possess appropriate rotational Jacobian singular values for quick tilting, yet it can avoid the singular and the special configurations within the stroke of each actuator. Actuator Radius $k_3=r_{actuator}/L$ was also chosen to meet other practical requirements: making room for the chosen actuators. In the end, our prototype Design Parameters were $k_1 = 0.2647$, $k_2 = 30^\circ$, $k_3 = 0.0186$ and $k_4 = 0.11765$.



\section{Conclusions} \label{conclude}
The Triple Scissor Extender is a viable choice of mobile manipulator for manufacturing and assembly in heavy industry because it combines a large height range with a small footprint. A TSE can be designed to extend from a low to a high area and perform any manipulation task in between while retracting fully for transportation in a compact and manageable package. Using the design parameter studies from this paper, a wide range of TSE robots can be designed to accomplish a variety of tasks with varying requirements.

\bibliographystyle{IEEEtran}
\bibliography{IEEEabrv,myBib}
	

\end{document}